\documentclass[lettersize,journal,twoside]{IEEEtran}
\usepackage{amsmath,amsfonts}
\usepackage{algorithmic}
\usepackage{algorithm}
\usepackage{array}
\usepackage[caption=false,font=normalsize,labelfont=sf,textfont=sf]{subfig}
\usepackage[table]{xcolor}  
\definecolor{lightgray}{rgb}{0.9, 0.9, 0.9}  
\usepackage{textcomp}
\usepackage{stfloats}
\usepackage{url}
\usepackage{verbatim}
\usepackage{graphicx}
\usepackage{cite}
\usepackage{makecell}
\usepackage{caption}            
\usepackage{soul}         
\sethlcolor{yellow}       
\usepackage{xcolor}
\usepackage{soulutf8}

\usepackage[colorlinks=true, linkcolor=blue, urlcolor=blue, citecolor=blue]{hyperref}
\soulregister\cite7
\soulregister\ref7

\begin{document}

\title{\textbf{II-NVM}: Enhancing Map Accuracy and Consistency \\ with \textbf{N}ormal \textbf{V}ector-Assisted \textbf{M}apping}

\author{Chengwei Zhao$^{*}$, Yixuan Li$^{*}$, Yina Jian, Jie Xu$^{\dag}$, Linji Wang, Yongxin Ma, Xinglai Jin
\thanks{Chengwei Zhao and Yixuan Li with the Institute of Artificial Intelligence and Robotics, Xi’an Jiaotong University, Xi’an 710049, China (e-mail: 2191312118@stu.xjtu.edu.cn).}
\thanks{Chengwei Zhao and Xinglai Jin are with Hangzhou Qisheng Intelligent Techology Co. Ltd., 4083 Jianshe Fourth Road, Hangzhou, 311217, Zhejiang, China (e-mail: chengweizhao0427@gmail.com; jinxinglai@qishengrobot.com).}%
\thanks{Yina Jian is with the Department of Computer Science, Columbia University in the City of New York, 116th and Broadway, New York, NY 10027, USA (e-mail: yj2713@columbia.edu). }
\thanks{Jie Xu, Linji Wang, Yongxin Ma are with the School of Electrical and Electronic Engineering, Nanyang Technological University, 639798, Singapore (e-mail: jeff\_xu\_0503@foxmail.com; lwang44@gmu.edu; yxma@mail.sdu.edu.cn).}
\thanks{$^{*}$Chengwei Zhao and Yixuan Li contributed equally to this work and should be considered co-first authors. $^{\dag}$Jie Xu is the corresponding author.}
\thanks{\textit{Code} — \href{https://github.com/chengwei0427/II-NVM}{https://github.com/chengwei0427/II-NVM} }
\thanks{\textit{Video} — \href{https://www.youtube.com/watch?v=qso39uI7l38}{https://www.youtube.com/watch?v=qso39uI7l38}}
\thanks{Digital Object Identifier (DOI): see top of this page.}}


\IEEEpubid{0000--0000/00\$00.00~\copyright~2021 IEEE}

\maketitle

\begin{abstract}
SLAM technology plays a crucial role in indoor mapping and localization. A common challenge in indoor environments is the ``double-sided mapping issue'', where closely positioned walls, doors, and other surfaces are mistakenly identified as a single plane, significantly hindering map accuracy and consistency. To addressing this issue this paper introduces a SLAM approach that ensures accurate mapping using normal vector consistency. We enhance the voxel map structure to store both point cloud data and normal vector information, enabling the system to evaluate consistency during nearest neighbor searches and map updates. This process distinguishes between the front and back sides of surfaces, preventing incorrect point-to-plane constraints. Moreover, we implement an adaptive radius KD-tree search method that dynamically adjusts the search radius based on the local density of the point cloud, thereby enhancing the accuracy of normal vector calculations. To further improve real-time performance and storage efficiency, we incorporate a Least Recently Used (LRU) cache strategy, which facilitates efficient incremental updates of the voxel map. The \href{https://github.com/chengwei0427/II-NVM}{code} is released as open-source and validated in both simulated environments and real indoor scenarios. Experimental results demonstrate that this approach effectively resolves the ``double-sided mapping issue'' and significantly improves mapping precision. Additionally, we have developed and open-sourced the first simulation and real-world dataset specifically tailored for the ``double-sided mapping issue''.

\end{abstract}

\begin{IEEEkeywords}
indoor SLAM, double-sided mapping issue, normal vector-assisted Mapping.
\end{IEEEkeywords}

\section{Introduction}

In SLAM tasks, systems are required to simultaneously perform pose estimation and map construction within unknown environments\cite{bresson2017simultaneous}. Accurate mapping is critical for robotic navigation and autonomous driving, particularly in complex indoor settings\cite{ismail2022exploration}. Additionally, the use of SLAM technology in indoor mapping is increasingly gaining attention due to its ability to provide high-precision spatial information, including local geometric details\cite{locus1}\cite{locus2}, which supports the digital management and maintenance of buildings and facilities\cite{digital}.

\begin{figure}
    \centering
    \includegraphics[width=1\linewidth]{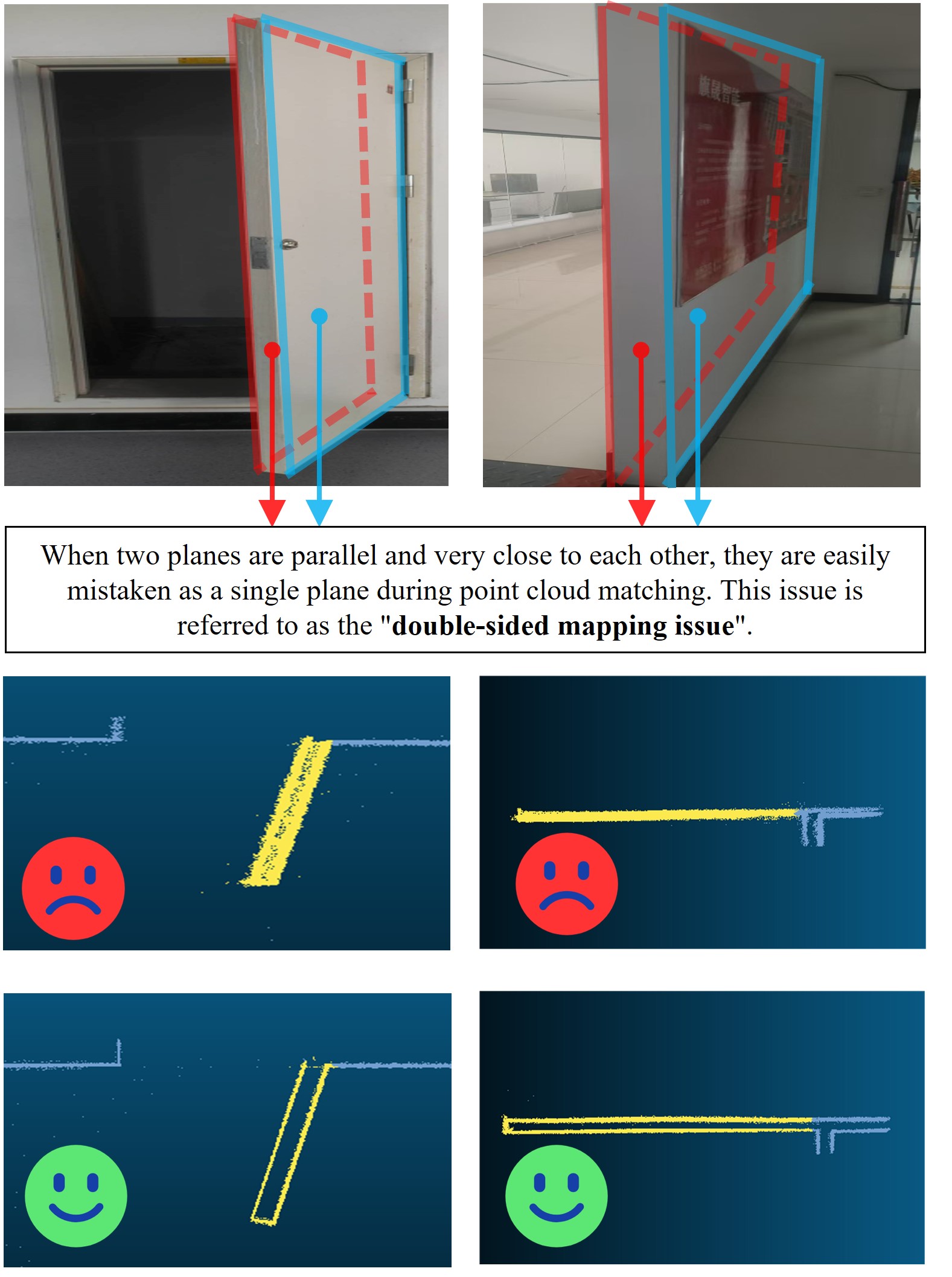}
    \caption{illustrates a real-world scenario where the double-sided mapping issue arises, presenting examples of both erroneous and correct mappings. }
    \label{fig:1}
\end{figure}

\IEEEpubidadjcol
The distinct structural features of indoor environments, including thin walls, doors, and windows, complicate LiDAR scan data analysis. Existing LiDAR-Inertial Odometry (LIO) algorithms, which are widely used outdoors, often underperform indoors\cite{adalio}\cite{log2}. Complex scenarios featuring multiple rooms and walls increase the difficulty of SLAM systems, as point clouds from both sides of thin walls may be erroneously matched as a single plane due to their proximity, causing the double-sided mapping issue (as illustrated in Fig. \ref{fig:1}). Many existing SLAM methods experience reduced mapping accuracy due to the lack of normal vector information or inadequate handling of normal vector consistency and map management.

To effectively resolve the double-sided mapping issue, this study introduces a method based on the calculation of point cloud plane normal vectors. This method is called II-NVM, where ``II'' looks like the double-sided wall, emphasizing its ability to accurately distinguish between the front and back sides of a wall during mapping. ``NVM'' stands for \textbf{N}ormal \textbf{V}ector-Assisted \textbf{M}apping, indicating that this method solves the double-sided mapping issue through normal vector consistency. This innovation markedly enhances the matching accuracy between point cloud data and object surface features, thereby improving the overall precision and consistency in mapping of map construction.

The main contributions of this paper include:
\begin{itemize}
  \item 
    Our paper enhances the storage of normal vectors within voxel maps, enabling each voxel block to not only contain point cloud data but also record normal vector information for both the front and back sides. This design facilitates the use of normal vector consistency in subsequent matching and map updates, effectively preventing the double-sided mapping issue. By incorporating this dual-sided storage approach, the method supports efficient incremental voxel map updates, ensuring real-time performance and improved storage efficiency.

\item We proposes a new adaptive radius KD-tree search method for calculating normal vectors, which dynamically adjusts the neighborhood search radius based on local point cloud density. This method analyzes the consistency of normal vector directions in point clouds to accurately distinguish the front and back of planar point clouds, effectively addressing mapping errors commonly encountered in traditional SLAM systems in double-sided scenarios.

\item We conducted both simulation and real-world experiments to verify the  effectiveness of the proposed method. To advance this research area, we have open-sourced the associated code and dataset, creating the first dataset specifically designed for the double-sided mapping issue, which provides a standardized evaluation benchmark and an important reference for future research.
\end{itemize}


\section{RELATED WORK}
\subsection{ Traditional LIO Mapping Algorithm }

LOAM\cite{zhang2014loam} is a foundational LiDAR SLAM system that facilitates pose estimation and map construction through geometric feature extraction and point cloud matching. Building on this, FAST-LIO\cite{xu2022fast} provides an efficient and robust LiDAR-Inertial Odometry solution, employing tightly coupled, iterative extended Kalman filtering (EKF) for real-time pose estimation and map construction, while significantly reducing computational demands. In a similar vein, I2EKF-LO\cite{i2ekf} utilizes dual iterative EKF to handle point cloud motion distortion, dynamically adjust process noise, and support diverse sensor platforms, achieving high-precision and efficient state estimation. For multimodal sensor fusion, M-DIVO\cite{xu2024m-divo} integrates visual, depth, and inertial modules from multiple ToF RGB-D cameras, utilizing an odometry system based on IEKF to enhance robustness, precision, and real-time performance through a multimodal redundancy scheduling mechanism and improved sensor calibration. In contrast, SuMa\cite{suma} employs a surface model-based mapping technique that uses dense point clouds to create accurate maps, while LIPS\cite{lips} leverages 3D indoor scenes by introducing nearest points to parameterize planes and utilizes a plane-to-plane cost for precise pose estimation.

Although geometric feature matching is commonly used in the aforementioned algorithms, planes are not effectively considered in double-sided areas. This can lead to pose errors and reduced accuracy in pose estimation. To address this limitation, we propose an adaptive radius KD-tree normal vector calculation method based on normal vector data to better manage the double-sided mapping issue.

\subsection{ Voxel Map-Based Mapping Method }
Faster-LIO \cite{faster} uses sparse voxel hash mapping for point cloud data storage, significantly reducing storage requirements while maintaining computational efficiency. In a similar vein, VoxelMap\cite{voxel} adapts to different environmental structures through an adaptive voxel size construction method, improving robustness for sparse and irregular LiDAR point clouds, and enhancing the efficiency of voxel construction, updating, and querying. 
CT-ICP \cite{cticp} further accelerates real-time processing by storing dense point cloud local maps within a sparse voxel framework.  

In contrast, our method utilizes normal vector information to address the double-sided mapping issue, employing incremental voxel updates and LRU cache strategies to manage point clouds more efficiently, thereby significantly enhancing real-time capabilities.

\subsection{ Normal Vector-Based SLAM Algorithm }
The LiDAR SLAM with Plane Adjustment method \cite{planeslam} resolves the double-sided mapping issue by calculating the plane normal vector \cite{normal} from the local surface geometry of the point cloud and aligning it towards the LiDAR center to differentiate the front and back of objects. Similarly, LOG-LIO \cite{log-lio} is a robust and precise LiDAR-Inertial Odometry system that emphasizes real-time estimation of the normal vectors of LiDAR scan points and the distribution of map points, ensuring effective utilization of these components. It also proposes a ring-based fast approximate least squares method for improved efficiency. On the other hand, NV-LIOM\cite{nvlio} extracts normal vectors from LiDAR scans and applies them in correspondence searches to improve point cloud registration. By analyzing the distribution of normal vector directions and checking for degeneracy, it adjusts correspondence uncertainty to enhance registration accuracy.

Existing normal vector-based SLAM methods face limitations in registration accuracy due to poor utilization and management of normal vectors. For example, LOG-LIO only uses normal vectors for field-of-view checks, while NV-LIOM stores normal vectors for each point and builds local maps from keyframes, but struggles with submap construction and normal vector inconsistencies. To address these issues, we propose an adaptive resolution normal vector estimation method that works with various LiDAR sensors. We also extend the voxelmap data structure to include normal vector data and enhance voxels to store both front and back side information, solving the double-sided mapping issue.

\section{METHOD}

\begin{figure*}[t]
    \centering
    \includegraphics[width=\textwidth]{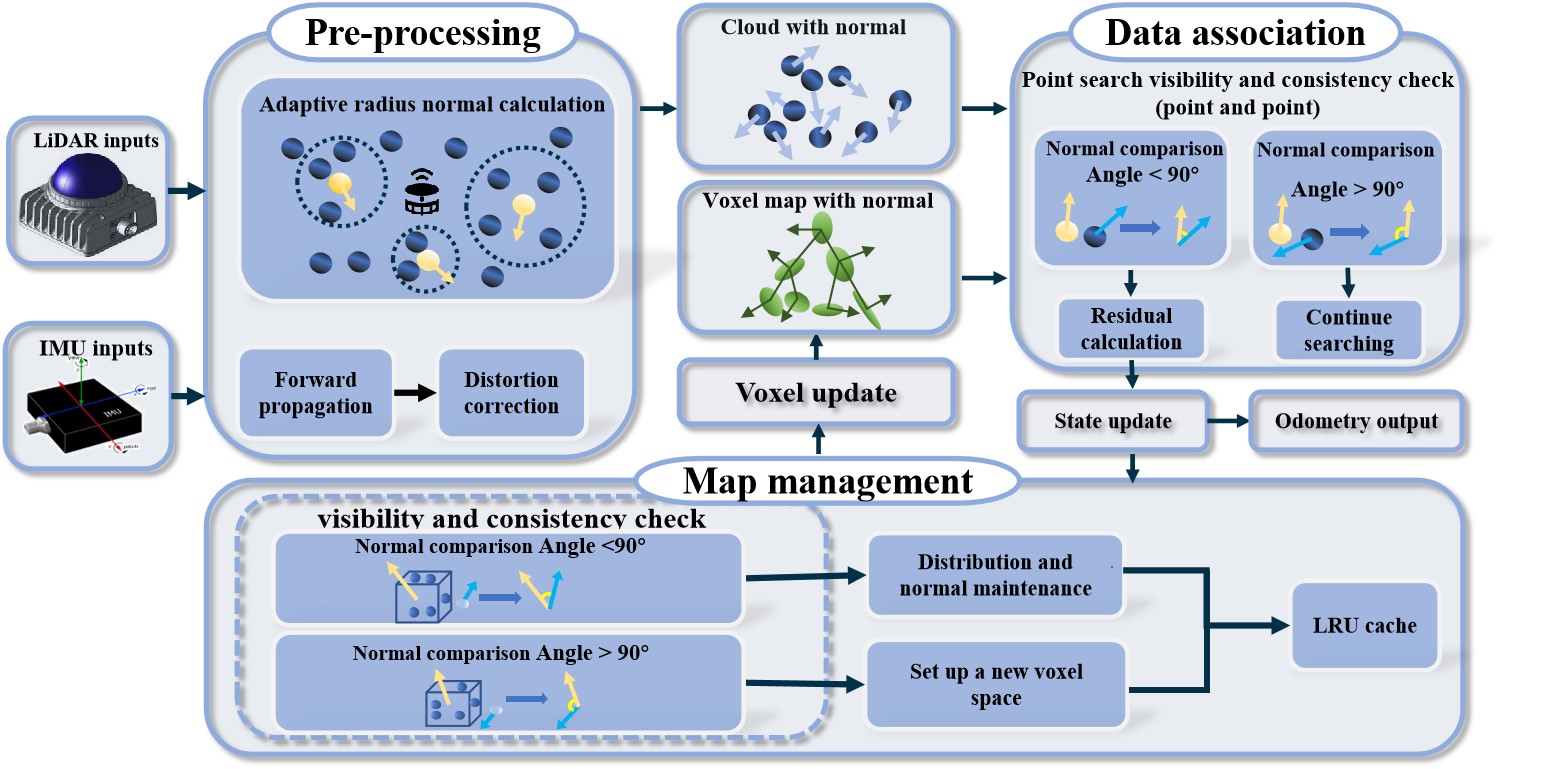} 
    \captionsetup{justification=centering} 
    \caption{System overview of II-NVM.}
    \label{fig:2}
\end{figure*}

The pipeline of II-NVM is shown in Fig. \ref{fig:2}. This section presents the core SLAM method we proposed, which aims to solve the ``double-sided mapping issue". Our method ensures the consistency of plane normal vectors and improves map accuracy through adaptive search radius normal vector calculation, incremental voxel map management, and normal vector view consistency processing.
\subsection{ Data Preprocessing and Normal Vector Calculation }

Accurately calculating normal vectors is crucial for maintaining consistency in plane normal vector discrimination. Traditional methods typically manage data using a KD-tree and employing a fixed-radius neighborhood search to estimate normal vectors. However, in complex indoor environments, the density of LiDAR point cloud data can vary significantly based on the openness of the environment and the distance from the LiDAR origin. The persistent use of a fixed-radius neighborhood search method struggles to adapt to the geometric features at varying distances, leading to decreased accuracy in normal vector calculations, especially in edge and corner areas. To address this, this paper proposes a distance-adaptive robust normal estimation method.

Due to the inaccuracy of LiDAR data for distance measurements in regions that are either too close or too far, we first filter out these point clouds. The filtered data is then organized using a KD-tree structure, enabling efficient iteration through the dataset to identify the nearest neighbors for each point. Next, the normal vector for each point is calculated using the eigenvalue decomposition method. To address the varying densities of point cloud data at different scanning radius, we introduce a distance-adaptive search radius, where the size of the search radius is set based on the scanning radius (Dist) of the current point, as shown in Equation (\ref{dist}):\\
\begin{equation}\label{dist}
r = \frac{D_{scan} - D_{\min}}{D_{\max} - D_{\min}} \cdot (R_{\max} - R_{\min}), 
\end{equation}
where $ R_{\max} $ and $ R_{\min} $ represent the maximum and minimum search radius, respectively. Similarly, $ D_{\max} $ and $ D_{\min} $ denote the maximum and minimum scanning radius and $ D_{\text{scan}} $ is the scanning distance of the current point (sensor-to-point range). 

Additionally, due to boundary discontinuities and multiple reflections, the covariance matrix decomposition is particularly sensitive to outliers. To address this, we employ two simple and effective methods for outlier elimination. Firstly, if the number of points in the nearest neighbor set is below a specified threshold \( Ng_{\text{thres}} \), the current point is considered an outlier. The second criterion is based on the planarity of the neighborhood set: assuming \( \lambda_1, \lambda_2, \lambda_3 \) are the eigenvalues of the plane covariance decomposition, arranged in descending order, and \( \delta_p \) represents its planarity. If \( \delta_p \) exceeds the threshold \( \delta_{\text{thres}} \), the current point is deemed an outlier. The planarity \( \delta_p \) is calculated as shown in Equation (\ref{planarity}):

\begin{equation}\label{planarity}
\delta_p = \frac{\lambda_1}{\lambda_1 + \lambda_2 + \lambda_3}.
\end{equation}Data identified as outliers will not be included in subsequent calculations.

\subsection{ Incremental Voxel Map }

We adopted a voxel map management strategy combined with an LRU caching strategy to achieve incremental voxel map construction. Each voxel block not only stores point cloud data but also records normal vector information to facilitates subsequent consistency checks of normal vectors.

\subsubsection{Intra-Voxel View Consistency Judgment} In each voxel block, point cloud data collected from different times and locations may have different normal vector distributions. Here, we describe how to judge view consistency. First, for data that needs to be inserted into a voxel, we calculate its view consistency with existing data. Assuming surface data exists within the current voxel, we determine whether the principal direction of the surface data has been calculated. If it has, we proceed to the next step, otherwise, we traverse all data on the current surface, extract all point normal vector information, construct a covariance matrix, and calculate the main direction of the normal vector distribution in the current voxel using covariance decomposition. 
Then, using the vector dot product formula, we compute the angle between the normal vector of the point to be inserted and the principal direction of the current voxel. If the angle is less than a threshold $\theta_{\mathrm{th}}$ (e.g., $90^{\circ}$), it is considered coplanar and the point is inserted into the current surface while updating the voxel's principal direction. If the angle exceeds $\theta_{\mathrm{th}}$, the point is stored on the opposite side of the current plane. Notably, $\theta_{\mathrm{th}}$ is empirically set to $90^{\circ} \pm 10^{\circ}$, where minor variations within this range do not compromise experimental robustness due to the voxel-based geometric tolerance.

\subsubsection{NVM Voxel Map Management} Traditional voxel map management involves dividing the environment into multiple voxel blocks and using hash values for accessing each block. Each voxel block stores the coordinate information of LiDAR point cloud data, reducing storage and computing complexity. To effectively resolve the double-sided mapping issue, we extended the traditional voxel map structure, allowing each voxel block to store not only coordinate information but also normal vector data for each point cloud. Additionally, considering that each object may have two planes (front and back), we expand each voxel block to store data for both planes. By default, data is stored on the front side first, and if view inconsistency is detected in the current voxel, the data is stored on the back side of the current plane. This extension enables the voxel block to store more geometric information, facilitating in precise associative matching in subsequent point correspondence searches and view consistency judgments.

Specifically, for incoming LiDAR data and corresponding global poses, we first transform point cloud data to the global coordinate system using Equation (\ref{transfer}), and calculate the hash index of the voxel map corresponding to the current point using Equation (\ref{hash}). When no voxel data exists for the current index, the current voxel block is initialized, and data is stored in it. If a voxel block corresponding to the current index already exists, the view consistency between the normal vector of the current point and the normal vectors of existing data in the voxel is evaluated. When the normal vector views are consistent, it indicates the point and existing points in the voxel block originate from the same surface, and the system updates the point cloud data into the front region of the voxel block. On the other hand, if normal vector views are inconsistent, suggesting the point and points stored in the voxel block may come from different sides of an object, to avoid the double-sided mapping issue, the system stores the point in the back region of the voxel block without resetting the entire voxel block.
\begin{equation}
\label{transfer}
P_{wi} = T P_{li}, \quad n_{wi} = R n_{li}, 
\end{equation}
where \( P_{wi} \) and \( n_{wi} \) are the position and normal vector in the global coordinate system, and \( P_{li} \) and \( n_{li} \) are the corresponding values in the local coordinate system. \( T \) is the translation matrix and \( R \) is the rotation matrix, which together describe the pose of the current point cloud.

\begin{equation}\label{hash}
v_i = \frac{P_{wi}}{d_i}, 
\end{equation}
where \( v_i \) represents the voxel index for the current point, \( P_{wi} \) is the position of the point in the global coordinate system, and \( d_i \) is is the resolution of the voxel map.

By dividing voxel blocks into front and back regions, the system can store data from different perspectives. The front region stores data consistent with the normal vector of the current observation point, while the back region is for data from the opposite surface. This approach allows a single voxel block to contain information from multiple views, effectively solving the double-sided mapping issue and preventing data loss and computational overhead associated with resetting voxel blocks. This method ensures that data stored in voxel blocks always reflects the correct geometric structure, preventing the double-sided mapping issue.

\subsubsection{LRU-Based Incremental Voxel Map Update} In managing incremental voxel maps, we introduce the LRU caching management strategy. LRU is a common memory management algorithm suitable for scenarios involving efficient handling of large data blocks. In SLAM, due to the vast amount of point cloud data, real-time map updates require effective voxel block management. It can lead to memory overflow or decreased system performance. By introducing the LRU management strategy, we can dynamically update and maintain voxel blocks based on the actual usage of point clouds. When a voxel block has not been accessed or updated for some time, it is marked as ``least recently used'' and is prioritized for replacement when storage space needs to be freed. This way, the system can ensure that storage resources are concentrated on the most active voxel blocks while reducing memory consumption. This mechanism not only enhances the efficiency of map updates but also provides a more flexible framework for handling normal vector consistency in map management.

\subsection{NVM-Based State Estimation}

To evaluate the effectiveness of our proposed algorithm, we integrated the NVM module into our open-source project, CT-LIO\footnote{\url{https://github.com/chengwei0427/ct-lio}}. The complete system workflow can be found in the project code, with this section focusing primarily on the data association process.

For data association, each point cloud datum is transformed into global coordinates using initial system-provided values. We calculate the hash index corresponding to each point using Equations (\ref{transfer}) and (\ref{hash}). This index is used to traverse the current index and its 26 neighboring indices.

\begin{figure}
    \centering
    \includegraphics[width=1\linewidth]{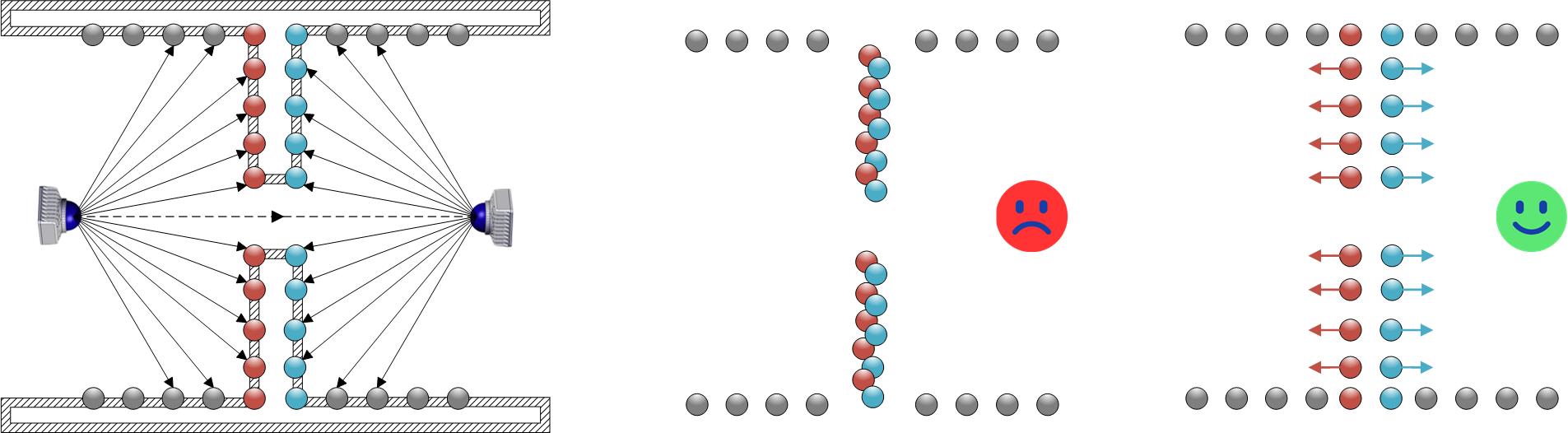}
    \caption{When the LiDAR scans through a wall, two closely spaced walls may be erroneously identified as a single plane. Accurate matching is achieved by employing normal vector consistency calculations.}
    \label{fig:3}
\end{figure}

We extract the data stored within the voxels corresponding to these indices. In addition to verifying that the distance from the current point meets specific criteria, we examine the angle between the normal vector of the target point and those of neighboring points to ensure they originate from the same surface. Similar to the Sec. III-B-1), if the angle between the normal vectors of two points is below the designated threshold, they are considered valid neighboring points. Specific details are illustrated in Fig. \ref{fig:3}.

After acquiring map data from neighboring points that satisfy the conditions, we construct the point-to-plane residual. Planar consistency is then applied to calculate the weight of the existing constraint, which is subsequently used to estimate the current state.

\section{EXPERIMENTAL RESULTS}

\subsection{ Description of Dataset and Experimental Environment }

In this study, directly comparing the performance of end-to-end SLAM methods presents certain challenges. One key issue is that some SLAM algorithms achieve high accuracy in an end-to-end setup without adequately addressing the double-sided mapping problem. Furthermore, obtaining ground truth data for indoor environments, particularly in large spaces spanning multiple rooms, is difficult. Traditional motion capture techniques are not capable of providing precise ground truth in such complex settings. As a result, conventional methods fall short in properly assessing how well SLAM systems handle the double-sided mapping issue.

To address these challenges, this study adopts two evaluation approaches to assess the performance of the proposed method. First, trajectory evaluation is performed using the Gazebo simulation environment, which generates precise ground truth trajectories, enabling accurate assessment of the algorithm's performance. Second, given the difficulty of obtaining accurate ground truth for real indoor datasets, we introduce wall thickness as an alternative evaluation metric. These two evaluation methods provide a comprehensive and objective means to validate the effectiveness of the proposed method in handling the double-sided mapping issue, ensuring both scientific rigor and fairness in the evaluation process. The experimental results are demonstrated in the video\footnote{\textit{Video} — \href{https://www.youtube.com/watch?v=qso39uI7l38}{https://www.youtube.com/watch?v=qso39uI7l38}} in the footnote.

Table \ref{dataset} summarizes the characteristics of the simulation datasets  used in this study, including the types of LiDAR, collection duration, trajectory length, and the number of poses. The datasets cover various environments, such as walls of different thicknesses, rooms, cafés, and garages.
\begin{table}[h]
    \centering
    \caption{\centering Simulation Dataset and LiDAR Specifications}
    \rowcolors{1}{white}{lightgray}  
    \renewcommand{\arraystretch}{1.5}
    \resizebox{\columnwidth}{!}{
    \Large
    \begin{tabular}{ccccc}
        \hline
        \textbf{Dataset Name} & \textbf{LiDAR Type} & \textbf{Duration (s)} & \textbf{Trajectory Length (m)} & \textbf{Pose} \\
        \hline
        \textit{Wall\_15cm\_a} & Livox & 149.092 & 54.673 & 4473 \\
        \textit{Wall\_15cm\_b} & Velodyne & 115.300 & 54.304 & 1154 \\
        \textit{Wall\_10cm\_a} & Livox & 240.441 & 69.761 & 7214 \\
        \textit{Wall\_10cm\_b} & Velodyne & 167.000 & 64.533 & 1671 \\
        \textit{Wall\_5cm\_a} & Livox & 248.480 & 48.425 & 7215 \\
        \textit{Wall\_5cm\_b} & Velodyne & 55.300 & 17.435 & 554 \\
        \textit{Room\_a} & Livox & 97.090 & 254.678 & 2913 \\
        \textit{Room\_b} & Velodyne & 115.800 & 50.675 & 1159 \\
        \textit{Café\_a} & Livox & 118.209 & 48.425 & 3546 \\
        \textit{Café\_b} & Velodyne & 90.900 & 42.632 & 910 \\
        \textit{Garage\_a} & Livox & 483.033 & 69.761 & 7214 \\
        \textit{Garage\_b} & Livox & 323.400 & 101.790 & 3235 \\
        \textit{Garage\_c} & Livox & 456.129 & 136.814 & 13684 \\
        \textit{Garage\_d} & Livox & 425.425 & 141.411 & 12763 \\
        \textit{Garage\_e} & Velodyne & 353.900 & 171.414 & 3540 \\
        \hline

    \end{tabular}
    }
    \label{dataset}
\end{table}

\begin{figure}
    \centering
    \includegraphics[width=1\linewidth]{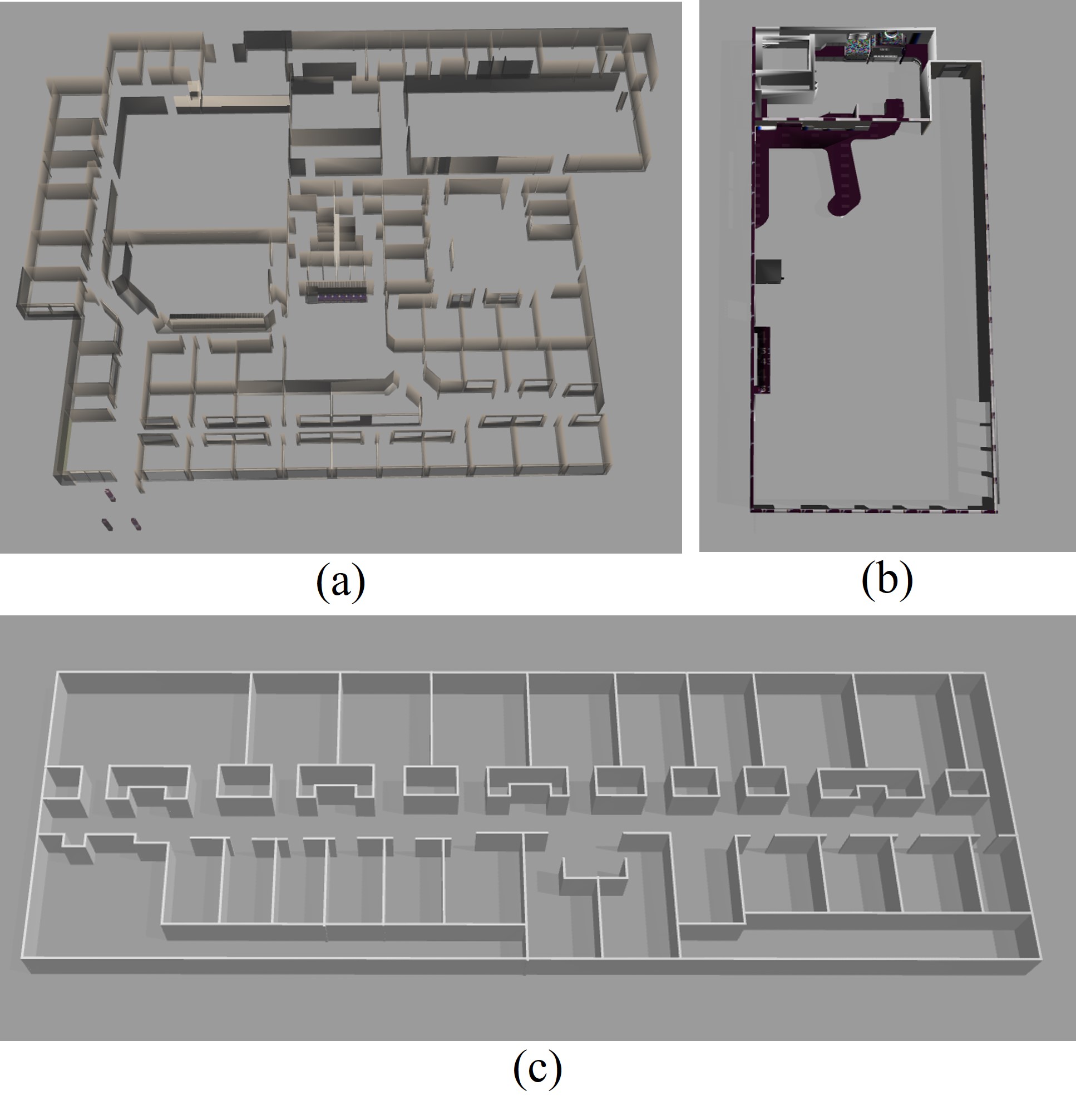}
    \caption{Gazebo simulation scenarios show (a) garage scene, (b) café scene, and (c) fixed-distance wall scene. }
    \label{fig:4}
\end{figure}

\subsection{Evaluation of Odometry in Gazebo Simulation Scenarios }In the Gazebo simulation environment, we built a complex indoor scene with various geometric structures, walls of different thicknesses, and double-sided areas, replicating real-world double-sided mapping challenges (Fig. \ref{fig:4}).To evaluate SLAM systems' pose estimation performance, we used Absolute Trajectory Error (ATE), which measures trajectory accuracy by comparing the estimated trajectory with the ground truth from the simulation. After running each method, we calculated and compared their ATE values \cite{grupp2017evo}.

Table \ref{ATElivox} and Table \ref{ATEvelodyne} present the quantitative results of Livox LiDAR and Velodyne LiDAR in various indoor environments.
In all our experiments, we employed a voxel size of 0.3m and successfully addressed the distortion issue that was absent in the simulation. CT-LIO and II-NVM (1m) demonstrated the trajectory accuracy without using normal vectors and the adaptive radius KD-tree.

\begin{table}[h]
    \centering
    \caption{\centering Results of Pose Estimation Comparison on Livox LiDAR }
    \renewcommand{\arraystretch}{2}
    \rowcolors{1}{white}{lightgray}  
    \resizebox{\columnwidth}{!}{
    \large
    \begin{tabular}{ccccccc} 
        \hline
        \textbf{Sequence} & \textbf{II-NVM} & \textbf{CT-LIO} & \textbf{II-NVM(1m)} & \textbf{FAST-LIO2} & \textbf{IG-LIO} & \textbf{DLIO} \\
        \hline
        \textit{Wall\_15cm\_a} & \textbf{0.0179} & 0.0236 & 0.0186 & 0.0545 & 0.0817 & 0.1161 \\
        \textit{Wall\_10cm\_a} & \textbf{0.0137} & 0.0155 & 0.0145 & 0.0492 & 0.0824 & 0.1145 \\
        \textit{Wall\_5cm\_a}  & \textbf{0.0163} & 0.1669 & 0.0168 & 0.0364 & 0.0766 & 0.0680 \\
        \textit{Room\_a}       & \textbf{0.0144} & 0.0150 & 0.0147 & 0.0570 & 0.0858 & 0.0729 \\
        \textit{Café\_a}       & \textbf{0.0168} & 0.0180 & 0.0171 & 0.0694 & 0.0846 & 0.1334 \\
        \textit{Garage\_a}     & \textbf{0.0162} & 0.0171 & 0.0165 & 0.0849 & 0.1096 & 0.1049 \\
        \textit{Garage\_b}     & \textbf{0.0158} & 0.0187 & 0.0188 & 0.0430 & 0.0446 & 0.0926 \\
        \textit{Garage\_c}     & \textbf{0.0133} & 0.0515 & 0.0135 & 0.1674 & 1.1304 & 0.1637 \\
        \textit{Garage\_d}     & \textbf{0.0138} & 0.0168 & 0.0147 & 1.3726 & 0.2051 & 0.1152 \\
        \hline
    \end{tabular}
    }
    \label{ATElivox}
\end{table}

\begin{figure*}[t] 
    \centering
    \includegraphics[width=\textwidth]{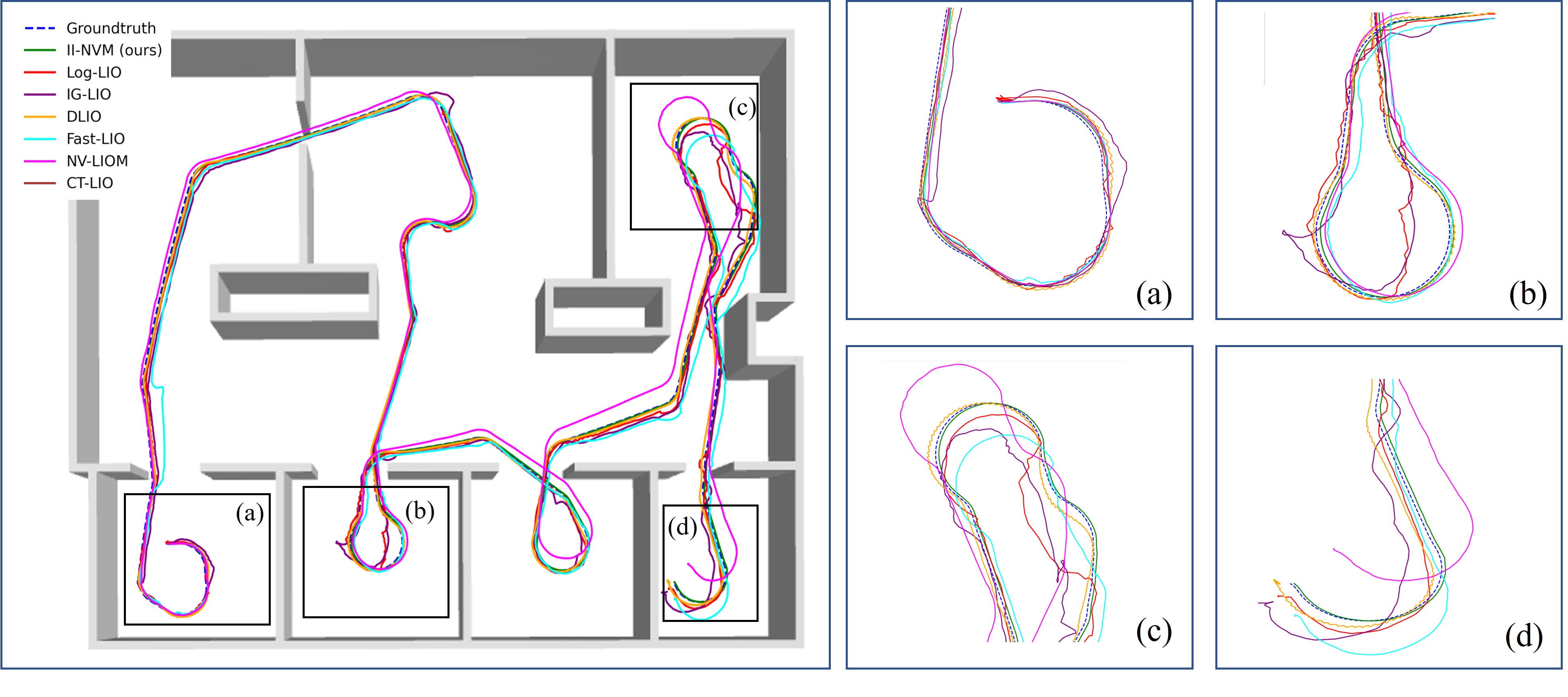} 
    \caption{Comparison of localization estimates from different algorithms.}
    \label{fig:odom_comparison}
    \label{fig:5}
\end{figure*}

Fig. \ref{fig:5} presents a comparison of trajectories from various algorithms for qualitative analysis. The experimental results demonstrate that in indoor environments with multiple walls, the II-NVM method significantly enhances trajectory estimation accuracy by effectively resolving double-sided mapping issues. It achieved the lowest ATE value, outperforming all other algorithms. 
Ablation experiments with CT-LIO and II-NVM (1m) further demonstrate the effectiveness of the adaptive radius KD-tree search method and the normal vector consistency check.
In contrast, the other methods showed larger errors in complex settings and struggled with the challenges posed by double-sided mapping. The findings highlight the superior mapping accuracy and reliability of the II-NVM method in multi-wall environments.

\begin{table}[h]
    \centering
    \caption{\centering Results of Pose Estimation Comparison on Velodyne LiDAR }
    \renewcommand{\arraystretch}{2}
    \resizebox{\columnwidth}{!}{
    \rowcolors{1}{white}{lightgray}  
    \LARGE
    \begin{tabular}{cccccccc} 
        \Xhline{0.6pt}
        \textbf{Sequence} & \textbf{II-NVM} & \textbf{CT-LIO} & \textbf{LOG-LIO} & \textbf{FAST-LIO2} & \textbf{IG-LIO} & \textbf{NV-LIOM} & \textbf{DLIO} \\
        \Xhline{0.6pt}
        \textit{Wall\_15cm\_b} & \textbf{0.0475} & 0.1466 & 0.3963 & 0.0686 & 0.2152 & 0.2344 & 0.0532 \\
        \textit{Wall\_10cm\_b} & \textbf{0.0474} & 0.0543 & 0.4026 & 0.0881 & 0.2030 & 0.3648 & 0.1145 \\
        \textit{Wall\_5cm\_b}  & \textbf{0.0410} & 0.0491 & 0.1693 & 0.1748 & 0.3067 & 0.1038 & 0.0986 \\
        \textit{Room\_b}       & \textbf{0.0641} & 0.1869 & 0.1078 & 0.0950 & 0.1506 & 0.1123 & 0.1083 \\
        \textit{Café\_b}       & \textbf{0.0206} & 0.0703 & 0.0930 & 0.1526 & 0.1260 & 0.0552 & 0.0958 \\
        \textit{Garage\_e}     & \textbf{0.2153} & 0.2460 & 0.6651 & 0.9372 & 0.6651 & 0.4602 & 0.2914 \\
        \Xhline{0.6pt}
    \end{tabular}
    }
    \label{ATEvelodyne}
\end{table}

\subsection{ Evaluation of Wall Thickness  }
\begin{figure}[h]
    \centering
    \includegraphics[width=1\linewidth]{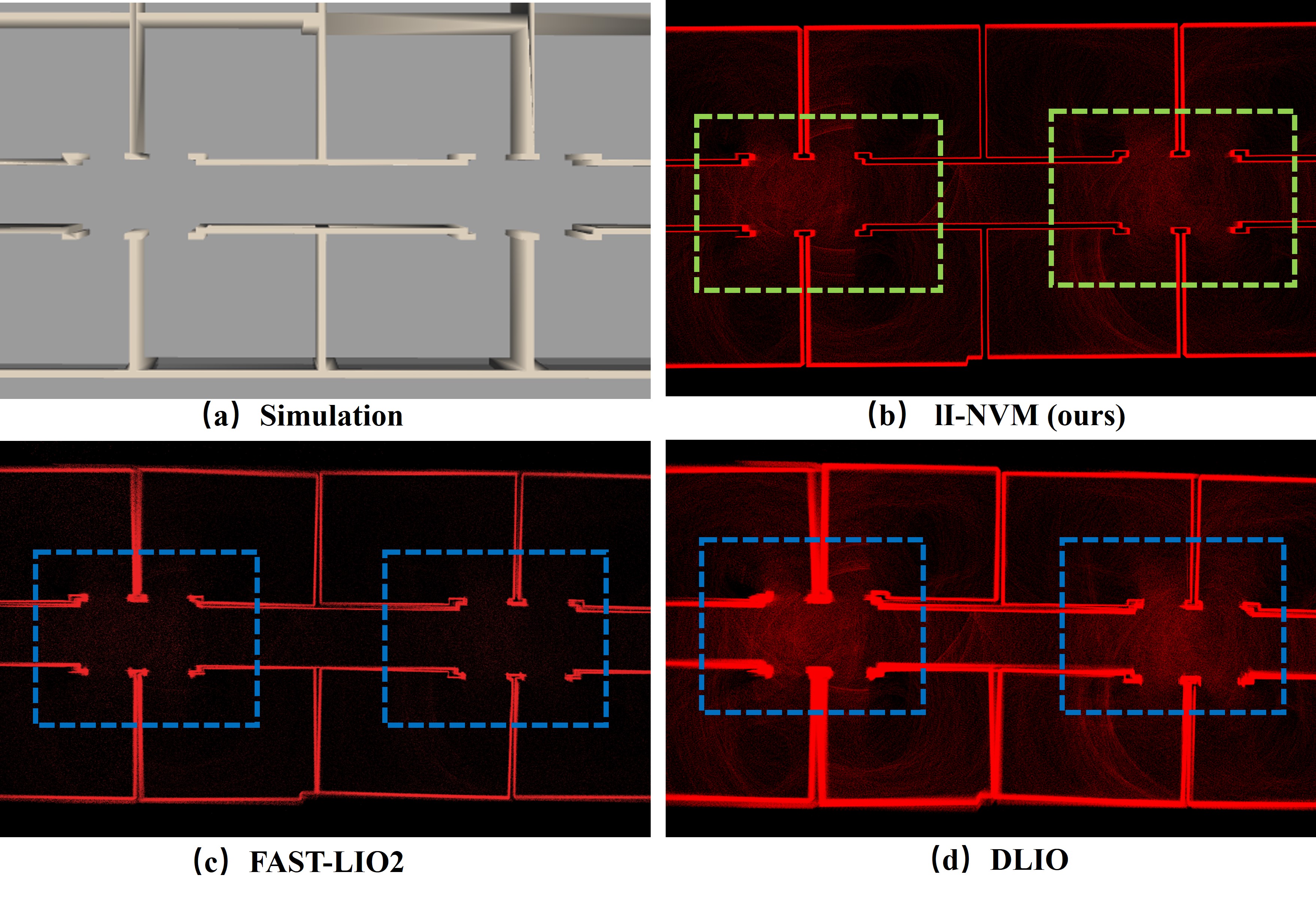}
    \caption{Mapping results of different algorithms in simulation.}
    \label{fig:6}
\end{figure}

We first compared the performance of various SLAM algorithms in addressing the double-sided mapping issues. The experimental results indicate that other SLAM algorithms have significant deficiencies when addressing double-sided mapping issues, as they are unable to effectively distinguish between the front and back sides of walls, leading to reduced mapping accuracy. In contrast, the II-NVM method successfully resolves the double-sided mapping issue, accurately identifying the front and back sides of walls and generating precise maps. The mapping results of different algorithms in the simulation environment are shown in Fig. \ref{fig:6}.

To validate the proposed method, we evaluated II-NVM using real and simulated data, focusing on its accuracy in restoring wall thickness in double-sided areas.

\begin{figure}[h]
    \centering
    \includegraphics[width=1\linewidth]{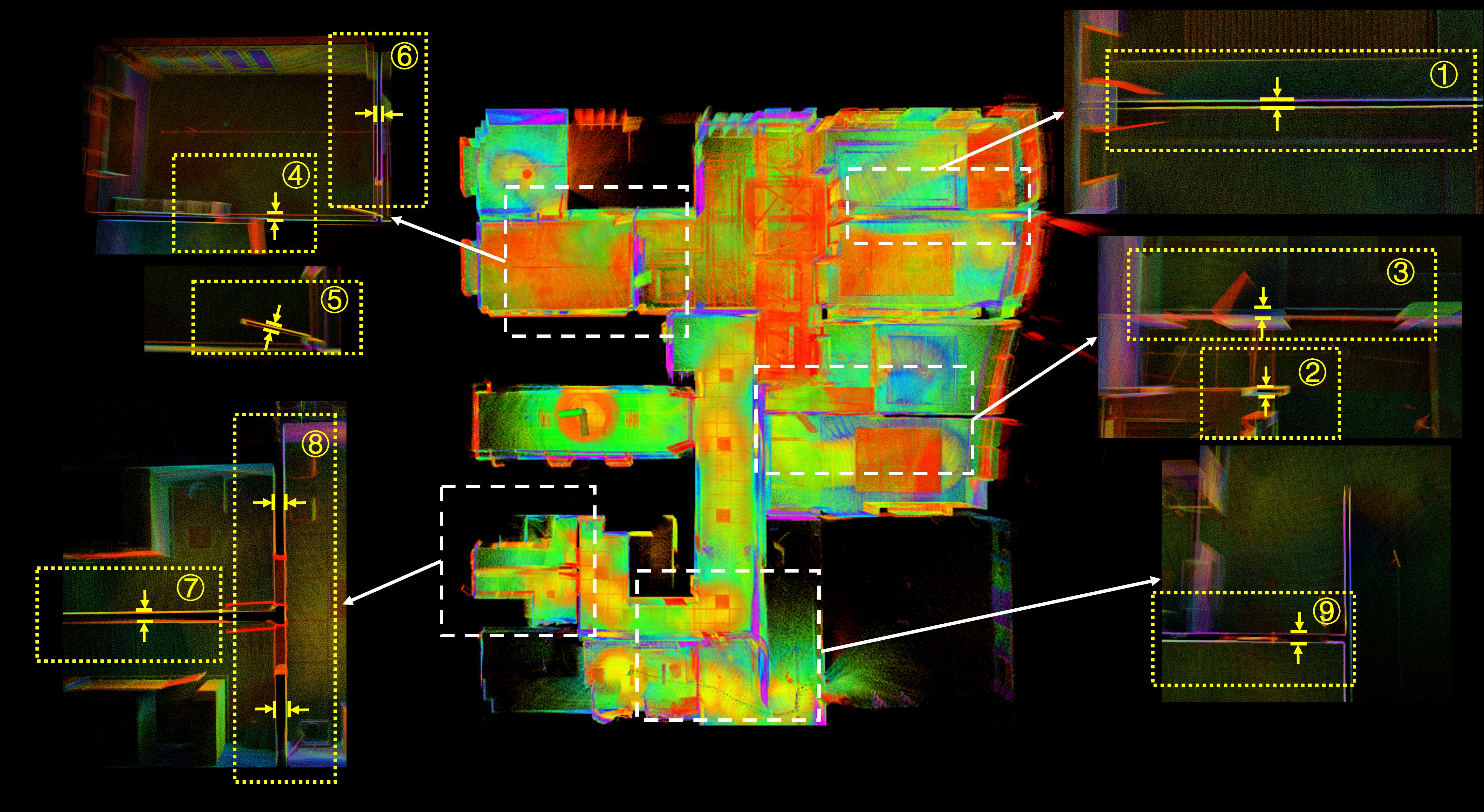}
    \caption{Mapping results of walls and doors in a real scenario.}
    \label{fig:7}
\end{figure}

In the experiment, point cloud data from indoor scenes was collected and processed using the II-NVM algorithm. To evaluate the algorithm's performance in double-sided areas, we extracted the double-sided regions from the point cloud after building the map and used a plane fitting method to determine the plane equation of one side of the wall. Based on the fitting results, the average distance between the plane of one side of the wall and the other was calculated as an evaluation metric for wall thickness, providing a quantitative assessment of the algorithm's accuracy in real scenarios. Since other SLAM algorithms cannot effectively address double-sided mapping issues and cannot measure wall thickness, this experiment further demonstrated the effectiveness of the proposed method in handling double-sided areas and restoring wall thickness through comparison with real point cloud data and some simulated point cloud data. Fig. \ref{fig:7} shows the mapping results in real scenarios, with numbered walls selected for further analysis and comparison.

\begin{table}[h]
    \centering
    \renewcommand{\arraystretch}{1.5}
    \rowcolors{1}{white}{lightgray}  
    \caption{\centering Comparison of II-NVM Double-Sided Mapping Performance Across Different Wall Thicknesses in Real-World Scenarios}
    \resizebox{1\columnwidth}{!}{
    \begin{tabular}{cccc}
        \hline
        \textbf{Area} & \textbf{Real Thickness (cm)} & \textbf{II-NVM (cm)} & \textbf{Percentage Change} \\
        \hline
        1 & 12.0 & 10.554 & -12.05\% \\
        2 & 12.0 & 12.242 & +2.02\% \\
        3 & 11.9 & 12.026 & +1.06\% \\
        4 & 11.9 & 11.652 & -2.08\% \\
        5 & 4.0 & 4.585 & +14.63\% \\
        6 & 9.1 & 9.550 & +4.95\% \\
        7 & 15.5 & 15.937 & +2.82\% \\
        8 & 15.9 & 14.737 & -7.32\% \\
        9 & 12.3 & 12.754 & +3.69\% \\
        \hline
    \end{tabular}
    }
    \label{real}
\end{table}

Since the simulation environment offers precise control over wall geometry and thickness, we conducted additional experiments to evaluate II-NVM’s performance across walls of varying thicknesses. By simulating seven wall areas with varying thicknesses, we further validated the adaptability and robustness of the proposed method under different conditions.

\begin{table}[h]
    \centering
    \caption{\centering Comparison of II-NVM Double-Sided Mapping Accuracy Across Different Wall Thicknesses in Simulation}
    \renewcommand{\arraystretch}{1.5}
    \rowcolors{1}{white}{lightgray}  
    \resizebox{1\columnwidth}{!}{
    \begin{tabular}{cccc}
        \hline
        \textbf{Area} & \textbf{Real Thickness (cm)} & \textbf{II-NVM (cm)} & \textbf{Percentage Change} \\
        \hline
        1 & 15.0 & 15.073 & +0.49\% \\
        2 & 13.0 & 13.127 & +0.98\% \\
        3 & 11.0 & 11.002 & +0.02\% \\
        4 & 9.0 & 9.161 & +1.79\% \\
        5 & 7.0 & 7.342 & +4.89\% \\
        6 & 5.0 & 5.032 & +0.64\% \\
        7 & 3.0 & 2.789 & -7.03\% \\
        \hline
    \end{tabular}
    }
    \label{simulate}
\end{table}

The experimental results, presented in Tables \ref{real} and \ref{simulate}, show the differences and relative errors between the actual wall thickness and the thickness estimated by the II-NVM method. Overall, the proposed method demonstrates high accuracy in estimating wall thickness in most scenarios, confirming its effectiveness in mapping double-sided areas. While some errors were observed in certain cases, the method exhibited good stability and reliability overall. In real-world wall thickness tests, the II-NVM method maintained measurement errors within 1\% for thicker targets (such as 15 cm and 13 cm), highlighting its excellent reliability. For thinner targets (such as 3 cm and 4 cm), although the error slightly increased, it remained below 10\%. These results further validate the effectiveness and practicality of the II-NVM method in addressing centimeter-level double-sided mapping challenges.

\subsection{ Evaluation of Processing Time }
By using a normal vector voxel map with an LRU cache, we ensure efficient response and updating of the voxel map, even in complex scenarios. This method enhances storage efficiency while maintaining real-time performance, providing stronger support for the widespread use of voxel maps in real-time applications. In this section, we will examine how the LRU cache module compares to the radius-based map management module (CT-LIO) in terms of time consumption for map updating, state estimation, pose optimization, and measurement processing for each sequence, as shown in the table.
\begin{table}[h]
    \centering
    \renewcommand{\arraystretch}{1.5}
    \rowcolors{1}{white}{lightgray}  
    \Large
    \caption{\centering Comparison of SLAM Operation Performance with and without LRU Cache}
    \resizebox{1\columnwidth}{!}{
    \begin{tabular}{cccc}
        \hline
        \textbf{Operation} & \textbf{II-NVM (s)} & \textbf{II-NVM\_w/o LRU  (s)} & \textbf{Percentage Change} \\
        \hline
        Map Update & 0.19 & 0.70 & +72.85\% \\
        Optimize & 3.34 & 4.60 & +27.39\% \\
        Pose Estimate & 3.58 & 5.29 & +32.33\% \\
        Process Measurement & 3.96 & 5.68 & +30.28\% \\
        \hline
        Total Time & 12.07 & 20.21 & +40.28\% \\
        \hline
    \end{tabular}
    }
    \label{LRU}
\end{table}

The evaluation of processing time, presented in Table \ref{LRU}, shows that integrating the normal vector voxel map with an LRU cache significantly enhances the processing efficiency of II-NVM and reduces overall time consumption. The processing time for all major operations was notably decreased, indicating that this method effectively enhances both the real-time performance and storage efficiency in complex scenarios, providing robust support for real-time applications.

\section{CONCLUSION AND FUTURE WORK}

The article addresses the prevalent challenge of double-sided mapping in SLAM systems, particularly in indoor scenarios, by proposing a solution anchored in normal vector consistency. This solution involves extending the storage of normal vector data within voxel blocks, alongside implementing an adaptive radius KD-tree search method and a view consistency determination mechanism, which effectively resolves the difficulty in distinguishing the front and back sides of objects. Experimental results demonstrate that this method significantly improves mapping accuracy in complex indoor environments and reduces double-sided mapping errors. Furthermore, by the integration of an LRU caching strategy enables efficient incremental updates of the voxel map, boosting real-time performance while maintaining accurate mapping.

Future research could develop this method as a standalone plugin, facilitating seamless integration with other LIO systems to further enhance the mapping accuracy of various SLAM systems in complex indoor environments.

\bibliographystyle{ieeetr}
\bibliography{1}

\end{document}